\documentclass[10pt,journal,compsoc]{IEEEtran}

\usepackage{epsfig}
\usepackage{graphicx}
\usepackage{amsmath}
\usepackage{amssymb}
\usepackage{multirow}
\usepackage{bm,makecell,amsthm}
\usepackage{url}
\usepackage{algorithm}
\usepackage{algorithmic}
\usepackage{array}
\usepackage{fixltx2e}
\usepackage{dblfloatfix}
\usepackage{mathrsfs}
\usepackage{url,color}

\begin{document}

\renewcommand{\algorithmicrequire}{\textbf{Input:}}   %
\renewcommand{\algorithmicensure}{\textbf{Output:}}  %

\title{Robust Guided Image Filtering}

\author{Wei~Liu, Xiaogang~Chen, Chunhua~Shen, Jingyi~Yu, Qiang~Wu and Jie~Yang

\IEEEcompsocitemizethanks{\IEEEcompsocthanksitem Wei Liu and Jie Yang are with the Key Laboratory of Ministry of Education for System Control and Information Processing, Shanghai Jiao Tong University, Shanghai, 200240, China. Email: \{liuwei.1989, jieyang\}@sjtu.edu.cn

\IEEEcompsocthanksitem Xiaogang Chen is with the College of Communication and Art Design, University of Shanghai for Science and Technology, Shanghai, 200093, China. Email:xg.chen@live.com

\IEEEcompsocthanksitem Chunhua Shen is with the School of Computer Science, University of Adelaide, Adelaide, SA 5005, Australia. Email: chunhua.shen@adelaide.edu.au

\IEEEcompsocthanksitem Jingyi Yu is with the College of Engineering, University of Delaware, Newark, DE 19176, USA. Email:yu@eecis.udel.edu

\IEEEcompsocthanksitem Qiang Wu is with the School of Computing and Communications, University of Technology, Sydney, NSW 2007, Australia. Email: qiang.wu@uts.edu.au}}

\IEEEtitleabstractindextext{%
\begin{abstract}
The process of using one image to guide the filtering process of another one is called Guided Image Filtering (GIF). The main challenge of GIF is the structure inconsistency between the guidance image and the target image. Besides, noise in the target image is also a challenging issue especially when it is heavy. In this paper, we propose a general framework for Robust Guided Image Filtering (RGIF), which contains a data term and a smoothness term, to solve the two issues mentioned above. The data term makes our model simultaneously denoise the target image and perform GIF which is robust against the heavy noise. The smoothness term is able to make use of the property of both the guidance image and the target image which is robust against the structure inconsistency. While the resulting model is highly non-convex, it can be solved through the proposed Iteratively Re-weighted Least Squares (IRLS) in an efficient manner. For challenging applications such as guided depth map upsampling, we further develop a data-driven parameter optimization scheme to properly determine the parameter in our model. This optimization scheme can help to preserve small structures and sharp depth edges even for a large upsampling factor ($8\times$ for example). Moreover, the specially designed structure of the data term and the smoothness term makes our model perform well in edge-preserving smoothing for single-image tasks (i.e., the guidance image is the target image itself). It performs well in several challenging applications in avoiding halos, gradient reversals and properly preserving edges with noise/texture being well smoothed. Through extensive experimental results, we show that the proposed RGIF can have promising performance in many applications such as guided depth map upsampling, flash/no flash filtering, detail enhancement, HDR tone mapping, structure smoothing and clip-art JPEG compression artifact removal.

{This paper is an extension of  our previous work \cite{liu2015data, liu2015robust}}.

\end{abstract}

}

\maketitle

\IEEEdisplaynontitleabstractindextext

\section{Introduction}
\label{SecIntroduction}

\IEEEPARstart{I}{mages} of low quality can suffer from noise, low resolution and other artifacts due to the light condition or the mechanism of the sensor. Color images captured under low light condition may contain noise and blurring edges. Due to the mechanism of Time-of-Flight (ToF) depth cameras, the obtained depth maps typically contain heavy noise and are of low resolution. Alternatively, images of high quality can be captured at the same position with better light condition or other kinds of sensors. For example, we can capture color images with much less noise and sharper edges by using flash light under low light condition. Color images of high resolution can be obtained with a camera when depth maps are captured. These high-quality images can be used as guidance to enhance the quality of those low-quality images. With the guidance of flash images, no flash images can be enhanced into images with much less noise and sharper edges \cite{agrawal2005removing, petschnigg2004digital, eisemann2004flash}. Color images of high-resolution can be used to guide the upsampling process of depth maps \cite{diebel2005application, yang2014depth_recovery, park2011high}. There are also other similar applications such as near infrared (NIR) image guided color image restoration \cite{krishnan2009dark, schaul2009color, zhang2008enhancing, zhuo2010enhancing}, day image guided night image enhancement \cite{raskar2005image} etc. Promising results were reported in these applications. In this paper, we denote such filtering process as Guided Image Filtering (GIF) which follows the idea proposed by He et al. \cite{he2013guided}. It is also called multispectral image restoration or cross-field image restoration \cite{shen2015multispectral, yan2013cross}.

\begin{figure*}
\centering
  \includegraphics[width=1\linewidth]{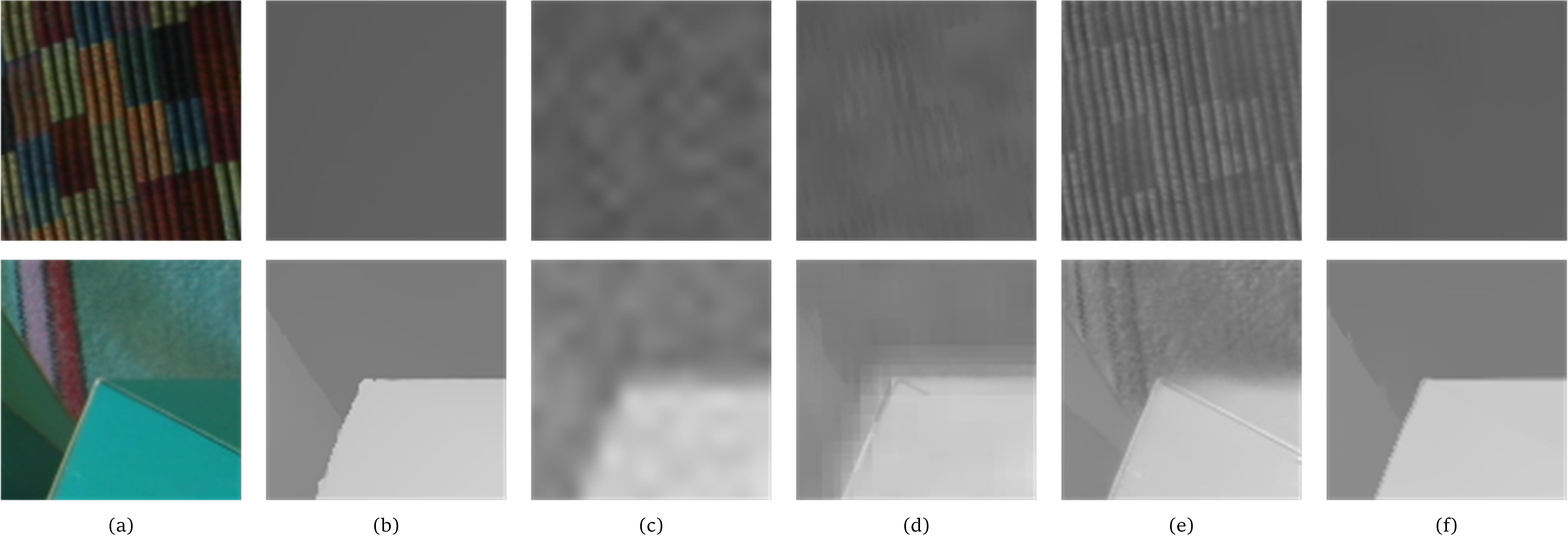}\\
  \caption{Illustration of texture copy and blurring edges cause by structure inconsistency in $8\times$ guided depth map upsampling. (a) Guidance images. (b) Groundtruth. (c) Target images (shown in bicubic interpolation) and filtered images by (d) JBU \cite{kopf2007joint}, (e) Shen et al. \cite{shen2015multispectral} and (f) ours.}\label{FigStructureInconsistency}
\end{figure*}

\textbf{The advantage of GIF} is that the guidance image can provide extra information especially the sharp edges that do not exist in the target image. The extra information makes the filtered image contain less noise but sharper edges.

\textbf{The challenge of GIF} is the structure inconsistency between the guidance image and the target image. The fundamental assumption of GIF is that both these two images have the same structures. When structures in these two images are consistent, there are seldom artifacts in the filtered image. We denote the situation where both two images have smooth regions or edges as \emph{structure consistency}. However, there are also situations where one image contains edges while the corresponding regions in the other image are smooth. We denote this as \emph{structure inconsistency} which may cause artifacts in the filtered image. In particular, artifacts are as follows:

\begin{itemize}
  \item \textbf{Texture copy} occurs when smooth regions in the target image correspond to highly textured regions in the guidance image. As a result, the corresponding regions in the filtered image contain similar structures as the guidance image has. Texture copy can cause notable artifacts in the filtered image in applications such as guided depth map upsampling \cite{liu2016variable} and flash/no flash image filtering \cite{shen2015multispectral}. We show one example of texture copy in $8\times$ guided depth map upsampling in the first row of Fig.~\ref{FigStructureInconsistency} .

  \item \textbf{Blurring edges} exist in the filtered image when edges in the target image correspond to smooth regions or quite weak edges in the guidance image. The second row of Fig.~\ref{FigStructureInconsistency} shows illustration.
\end{itemize}

In some applications such as guided depth map upsampling, input depth maps captured by ToF depth cameras usually contain quite \textbf{heavy noise}. How to handle the heavy noise is also a challenging issue as strong noise smoothing could also smooth out depth edges and small structures.

\textbf{How to handle the challenge of GIF} has been studied in many related methods. Shen et al. \cite{shen2015multispectral} used an additional shadow detection method to avoid the blurring edges caused by the shadow of NIR images in guiding the restoration of color images. Similar methods were also adopted by Petschnigg et al. \cite{petschnigg2004digital} and Eisemann et al. \cite{eisemann2004flash} for shadow detection in flash images. Yang et al. \cite{yang2014depth_recovery} proposed to combine the bicubic interpolation of input depth maps and guidance color images to avoid texture copy artifacts and blurring depth edges. To handle the heavy noise on depth maps, Park et al. \cite{park2011high} proposed to firstly use an MRF \cite{boykov2001fast} to clean up the noisy depth values on input depth maps and then perform GIF. However, these methods either are only suitable to a special application or have certain limitations.

In this paper, we propose a novel technique for Robust Guided Image Filtering (RGIF). We attempt to solve the above challenging issues in GIF with an optimization framework that consists of a data term and a smoothness term. Moreover, our model works well in handling several challenging single-image tasks (i.e., the guidance image is the target image itself). The main contributions of this paper are as follows.
\begin{itemize}

  \item[1.] Unlike previous methods \cite{petschnigg2004digital, eisemann2004flash, shen2015multispectral}  that adopted additional methods to handle the structure inconsistency. Our model itself can well handle the structure inconsistency. The key idea in our model is to make use of the property of both the guidance image and the target image through a novel smoothness term. Instead of separating smoothing the noise in target images and performing GIF as two steps like the work proposed by Park et al. \cite{park2011high}, our data term accommodates them simultaneously. We show that this makes our model have better performance. Moreover, both the special structure of the data term and the smoothness term make our model capable of handling many challenging single-image tasks.

  \item[2.] While the proposed model is highly non-convex, we propose a numeric solution with Iteratively Re-weighted Least Squares (IRLS) such that the problem can be efficiently solved by iteratively solving a linear system using off-the-shelf linear system solvers.

  \item[3.] For challenging applications such as guided depth map upsampling, we present a data-driven optimization scheme to properly determine the parameter in our model. This parameter optimization scheme helps to preserve small structures and sharp depth edges even for a large upsampling factor such as $8\times$.

  \item[4.] Our model outperforms many state-of-the-art methods in both dual-image tasks and single-image tasks.

 \end{itemize}

The rest of this paper is organized as follows: in Sec.~\ref{SecRelatedWork}, we show the related work of this paper. We then present our work in Sec.~\ref{SecRGIF} together with its numeric solution and analysis. A data driven parameter optimization approach is presented to properly adapt the parameter in our model. We also show that several models are related to our model as a special case of our model. Applications, their corresponding experimental results and comparison with other state-of-the-art methods are shown in Sec.~\ref{SecAppResults}. We draw the conclusion of our work in Sec.~\ref{SecConclusion}.

\begin{figure*}
\centering
  \includegraphics[width=1\linewidth]{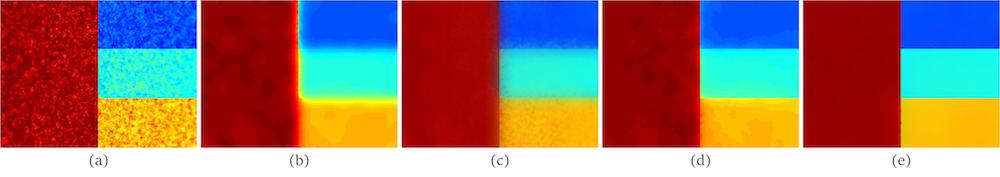}\\
  \caption{Visual comparison of edge preserving and noise smoothing. (a) Color visualized noisy input. Result of (b) BF \cite{tomasi1998bilateral}, (c) GF \cite{he2013guided}, (c) WLS \cite{farbman2008edge} and (d) our method.}\label{FigNoiseSmoothingIllustration}
\end{figure*}
\section{Related Work}
\label{SecRelatedWork}

Several methods are related to our work. We roughly classify these methods as \emph{local methods or filter-based methods} and \emph{global methods}. Local methods are based on certain edge-preserving filters where each pixel in the filtered image is a weighted sum of its neighbors in the target image. In GIF, the weight is usually based on guidance images. As a variant of Bilateral Filtering (BF) \cite{tomasi1998bilateral}, Joint Bilateral Filtering (JBF) has been widely used in many applications such as flash/no flash image filtering \cite{petschnigg2004digital}, joint image upsampling \cite{kopf2007joint}, to name a few. He et al. \cite{he2013guided} proposed a new edge-preserving filter named Guide Filter (GF). Due to its computational efficiency, GF has also been widely used in many applications such as stereo matching \cite{rhemann2011fast}, image matting \cite{he2010fast}, single image dehazing \cite{he2013guided}, etc. GF was further improved to preserve sharp image edges \cite{lu2012cross} and spatial variation on depth maps \cite{tan2014multipoint}. Inspired by the tree filter in image smoothing \cite{bao2014tree} and stereo matching \cite{yang2015stereo}, Dai et al. \cite{dai2015fully} extended GF \cite{he2013guided} to fully connected GF. Local methods are quite limited in smoothing the heavy noise in target images. When performing GIF, they also cannot well handle the structure inconsistency since their guidance weights are only based on guidance images. Recently, based on GF \cite{he2013guided}, Shen et al. \cite{shen2015mutual} proposed mutual-structure joint filtering which considered the structure inconsistency and promising results were shown.

Global methods usually formulate the filtering process as an optimization framework which takes the global property of an image into account. The filtered image is the global/local minimum of the objective function. The Weighted Least Squares (WLS) model is a basic optimization framework that was originally used for image debluring \cite{lagendijk1988regularized}. Recently, it has also been used in multi-scale tone mapping and detail manipulation \cite{farbman2008edge}, stereo matching \cite{min2008cost} and guided depth map upsampling \cite{min2014fast}. Take guided depth map upsampling for example, when performing GIF, WLS also suffers from the problem of texture copy and blurring edges because the guidance weight is completely based on the guidance image. Shen et al. \cite{shen2015multispectral} proposed to perform GIF with an optimized scale map. The assumption of their work is that the gradient of the filtered image is related to the scaled gradient of the guidance image. The limitation of such assumption is that it can cause blurring edges in the filtered image when there are no edges or quite weak edges in the guidance image corresponding to edges in the target image. This is validated by the experimental results in both their paper \cite{shen2015multispectral} and the one illustrated in Fig.~\ref{FigStructureInconsistency} (e).

Our work is motivated by the guided depth map upsampling framework proposed by Yang et al. \cite{yang2014depth_recovery} . To handle the structure inconsistency, Yang et al. \cite{yang2014depth_recovery} proposed to make use of not only the guidance color image but also the bicubic interpolation of the input depth map for the guidance weight. Promising results were shown in suppressing texture copy artifacts and preserving depth edges. In fact, the mechanism behind this idea is to make use of the property of both the guidance image and the depth map itself. This is because the bicubic interpolation can be considered as an approximation of the output depth map especially for small upsampling factors (e.g. $2\times$ and $4\times$). However, their method is not suitable for large upsampling factors because the bicubic interpolation becomes unreliable especially when the upsampling factor is large (e.g., $8\times$) and the input depth map contains heavy noise. Instead, when applied to guided depth map upsampling, our model can iteratively exploit the newly updated depth map in solving the optimization framework. The quality of the newly updated depth map is considerably better than that of the bicubic interpolation of the input depth map. This means we can make better use of the property of depth maps. This makes our model have better performance in suppressing texture copy artifacts and preserving sharp depth edges.

Due to the special structure of the data term and the smoothness term of our model, it is also an edge-preserving smoother with strong noise smoothing ability. Fig.~\ref{FigNoiseSmoothingIllustration} shows comparison of our method with other local and global methods in strong noise smoothing and edge preserving filtering. We will show in the experimental part that the nice edge-preserving smoothing property of our model make it capable of several challenging single-image tasks.

Some preliminary work of the proposed method appeared in \cite{liu2017robust}. However, in this paper we have substantially further developed the method in \cite{liu2017robust} and extended the results as follows. First and most importantly, the model in this paper is a general framework that is not limited to the problem of guided depth map restoration and can be applied to many dual-image tasks as well as single-image tasks. In contrast, only depth map restoration was considered in \cite{liu2017robust}, and the method there is tailored for depth map restoration. Second, our data term is robust against the noise while the data term in \cite{liu2017robust} is not. In addition, the special structure of our data term makes our model capable of coping with challenging single-image tasks such as texture smoothing \cite{xu2012structure} while the method in \cite{liu2017robust} does not perform well for such tasks. Third, the numeric solution proposed in this paper here is more efficient and the convergence is guaranteed. However, the convergence of the numeric solution proposed in \cite{liu2017robust} is not guaranteed. Fourth, we adapt the parameter in our model with an optimization framework while theirs is empirically determined. The last one is that we show several existing methods are special cases of our model in this paper. This was not explored in \cite{liu2017robust}. Moreover, we are the first to show the relation between the adopted error norm function and the $L_2$ error norm in this paper which was not explored in the literature.

\section{Robust Guide Image Filtering}
\label{SecRGIF}

\subsection{The Model}
\label{SecModel}
The inputs of our model are a target image to be filtered which is denoted as $I^0$ and a guidance image which is denoted as $G$. Both $I^0$ and $G$ can be single-channel images or multi-channel images depending on the application. If $I^0$ is a multi-channel image, we process each channel separately as adopted by Shen et al. \cite{shen2015multispectral, shen2015mutual}.

Our model consists of two terms: the data term and the smoothness term. Given $I^0$ and $G$, our model is formulated as:
\begin{equation}\label{EqOurModel}
\small{
    I^\ast=\underset{I}{\mathop{\arg }}\min\,\left\{( 1-\alpha)E_D(I,I^0)+\alpha E_S(I)\right\}
    }
\end{equation}
where $E_D(I, I^0)$ is the data term that makes the result to be consistent with the input target image. $E_S(I)$ is the smoothness term that reflects prior knowledge of the smoothness of our solution. The relative importance of these two terms is balanced with the parameter $\alpha $. The output of our model (i.e., the filtered image) is denoted as $I^\ast$.\\
\textbf{The data term $E_D(I, I^0)$} is defined as:
\begin{equation}\label{EqDataTerm}
\small{
    E_D(I,I^0)=\sum\limits_{i\in \Omega }\sum\limits_{j\in N_D(i)}\omega_{i,j}\varphi_D(|I_i-I_j^0|^2)
    }
\end{equation}
where $\Omega$ represents all the coordinates of the target image. $N_D(i)$ is the neighborhood of pixel $i$ which is a $(2r_d+1)\times(2r_d+1)$ square patch centered at $i$. The Gaussian window $\omega_{i,j}$ decreases the weights when $j$ is far from $i$:
\begin{equation}\label{EqSpatialWeight}
\small{
    \omega_{i,j}=\exp \left( -\frac{|i-j{{|}^{2}}}{2\sigma _{d}^{2}} \right)
}
\end{equation}
where $\sigma_d$ is a parameter defined by the user. ${\varphi}_{D}(\cdot)$ is the robust error norm function that we denote as \emph{exponential error norm}:
\begin{equation}\label{EqErroNormFunction}
\small{
    {{\varphi}_{D}}(x^2)=2\lambda^2\left(1-\exp \left(-\frac{x^2}{2\lambda^2} \right) \right)
    }
\end{equation}
where $\lambda$ is a user defined constant.

The design of our data term is robust against the noise in target images. This is implemented with two parts: the pixel-to-patch difference and the exponential error norm. The idea of pixel-to-patch difference has long been exploited in applications such as image denoising \cite{mrazek2006robust} and image editing \cite{an2008appprop}. This kind of data term is also called aggregated data term \cite{min2014fast}. It has been shown to be robust against noisy input images (for image denoising \cite{mrazek2006robust}) and inaccurate input strokes (for image editing \cite{min2014fast, an2008appprop}). Since the target image in our model can be noisy such as the depth map in guided depth map upsampling and the color image in flash/no flash image filtering, we thus also employ the similar pixel-to-patch difference measurement in the data term of our model. However, the shortage of such pixel-to-patch difference is that it may blur edges. To better preserve edges, we adopt the robust error norm function defined in Eq.~(\ref{EqErroNormFunction}) to model the data term. The adopted exponential error norm is known to be robust against outliers \cite{pizarro2010generalised} and thus could better preserve edges. The Gaussian window in Eq.~(\ref{EqSpatialWeight}) is introduced to further reduce the influence of pixels far from the central pixel. In addition, the proposed data term also benefit other tasks such as texture smoothing \cite{xu2012structure} other than handling the noise in target images. We will show more analysis and mathematical explanation in Sec.~\ref{SecAnalysis}.\\
\textbf{The smoothness term $E_S(I)$} should be robust against the structure inconsistency between the guidance image and the target image. This is achieved with:
\begin{equation}\label{EqSmoothnessTerm}
\small{
E_S(I) =\sum\limits_{i\in \Omega}\sum\limits_{j\in N_S(i)}\omega_{i,j}^g\varphi_S\left(|I_i-I_j|^2\right).
}
\end{equation}
where $N_S(i)$ is the neighborhood of pixel $i$ which is a $(2r_s+1)\times(2r_s+1)$ square patch centered at $i$.

The guidance weight $\omega_{i,j}^g$ is defined as:
\begin{equation}\label{EqColorSpatialWeight}
\small{
    \omega_{i, j}^g=\exp \left( -\frac{|i-j|^2}{2\sigma_{s}^2} \right)\cdot \exp\left(-\frac{\sum\limits_{k\in C}{|G_i^k-G_j^k|^2}}{|C|\times2\sigma _g^2}\right)
    }
\end{equation}
where $C$ represents different channels of $G$ which can be multiple channels or single channel depending on the application. $|C|$ represents the number of channels in $C$. $\sigma_s$ and $\sigma_g$ are parameters defined by the user.

We also employ the function in Eq.~(\ref{EqErroNormFunction}) to model the smoothness term:
\begin{equation}\label{EqSoomthnessNormFun}
\small{
    \varphi_S(\cdot)=\varphi_D(\cdot)
    }
\end{equation}

The guidance weight in Eq.~(\ref{EqColorSpatialWeight}) and the exponential error norm in Eq.~(\ref{EqSoomthnessNormFun}) help our smoothness term to make use of the property of both the guidance image and the target image, which is the key idea of our model to handle the structure inconsistency. We will present detailed mathematical explanation in Sec.~\ref{SecAnalysis}.

\subsection{Numeric Solution}
\label{SecNumSolution}

\begin{figure*}
\centering
  \includegraphics[width=1\linewidth]{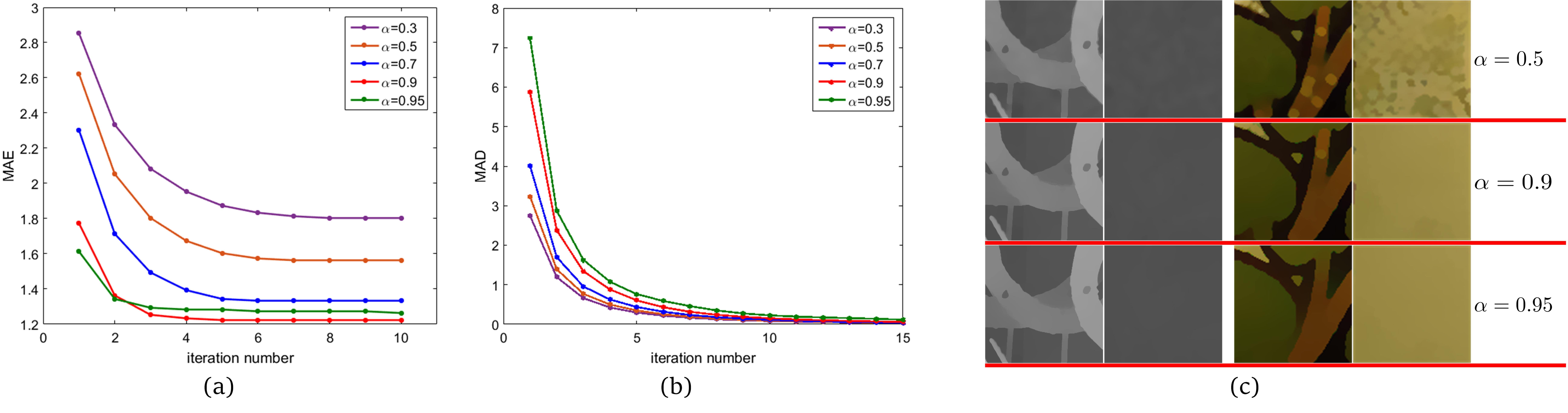}\\
  \caption{Convergence analysis of the proposed numeric solution. (a) MAE between the newly updated depth map and the groundtruth in each iteration on $8\times$ guided depth map upsampling for different $\alpha$ values. (b) MAD between adjacent two iterations on texture smoothing for different $\alpha$ values. (c) Examples of $8\times$ guided depth map upsampling and texture smoothing for different $\alpha$ values.}\label{FigCovergeAnalysis}
\end{figure*}

Due to the highly non-convex property of the proposed model, directly solving it is challenging. Previous work such as the one proposed by Lu et al. \cite{lu2011revisit} used Loop Belief Propagation (LBP) \cite{yedidia2000generalized} to solve their energy minimization function. However, classical energy minimization solvers such as LBP \cite{yedidia2000generalized} and graph cuts \cite{boykov2001fast, kolmogorov2004energy} work for discrete energy minimization. For most applications in this paper, the variables are naturally real-valued and one has to discretize them in order to apply LBP or graph cuts. The computational cost can be extremely expensive when the continuous problem is discretized into multiple levels. Usually, there are $256$ levels in natural color images and even more levels for depth maps captured by modern depth cameras which can be $8192$ levels ($13$ bits). The quantization error can be large if only a few quantization levels are used. In contrast, we present a numeric solution to our model that works for continuous system and can be efficiently solved. First, we present the normal equation of our model:
\begin{equation}\label{EqNormalEquation}
\small{
\begin{split}
    &\frac{\partial{E}}{\partial{I_i}}=(1-\alpha)\sum\limits_{j\in N_D(i)}\omega_{i,j}d_{i,j}\left(I_i-I_j^0\right)\\
    &\ \ \ \ \ \ \ \ \ \ \ \ \ \ \ \ \ +2\alpha\sum\limits_{j\in N_S(i)}\omega_{i,j}^gs_{i,j}\left(I_i-I_j\right) = 0, \ i \in \Omega\\
\end{split}
}
\end{equation}
where we define
\begin{equation}\label{EqErrorNormFunctionDerivative}
\small{
\begin{split}
    &d_{i,j}={\varphi}'_{D}(|I_i - I^0_j|^2), \ \ \ s_{i,j}={\varphi}'_{S}(|I_i - I_j|^2),\\
    &\ \ \ \ \ \ \ \ \ \ \ \ \ {\varphi}'_{D}(x^2)={\varphi}'_{S}(x^2)=\exp\left(-\frac{x^2}{2\lambda^2}\right)
\end{split}
}
\end{equation}
${\varphi}'_{D}(x^2)={\varphi}'_{S}(x^2)$ is the derivative of $\varphi_D(x^2)=\varphi_S(x^2)$ defined in Eq.~(\ref{EqErroNormFunction}).

A closed-form solution to Eq.~(\ref{EqNormalEquation}) is not available. We can only solve it iteratively. If we keep $s_{i,j}, d_{i,j}$ as constant in each iteration where $d_{i,j}^n={\varphi}'_{D}(|I_i^n - I^0_j|^2), s_{i,j}^n={\varphi}'_{S}(|I_i^n - I^n_j|^2)$ for iteration $n+1$, then Eq.~(\ref{EqNormalEquation}) becomes the standard form of the  following re-weighted least squares optimization framework as:
\begin{equation}\label{EqIterReWeightLeastSquare}
\small{
\begin{split}
    &I^{n+1}=\underset{I}{\arg }\min\{(1-\alpha)\sum\limits_{i\in\Omega}\sum\limits_{j\in N_D(i)}\omega_{i,j}d_{i,j}^n(I_i-I_j^0)^2 \\
    &\ \ \ \ \ \ \ \ \ \ \ \ \ \ \ \ \ \ \ \ \ \ \ \ \ \ \ \ \ \ \ \ \ \ \ \ + \alpha\sum\limits_{i\in\Omega}\sum\limits_{j\in N_S(i)}\omega_{i,j}^{g}s_{i,j}^n(I_i-I_j)^2\}
\end{split}
}
\end{equation}

Then we can iteratively solve Eq.~(\ref{EqIterReWeightLeastSquare}) until the final output meets the convergence condition. This approximation is similar to the well-known Iteratively Re-weighted Least Squares (IRLS) \cite{chartrand2008iteratively} in the literature. However, their IRLS is only suitable for $L_p (0<p<2)$ norm optimization framework. In this paper, we also denote Eq.~(\ref{EqIterReWeightLeastSquare}) as IRLS. As Eq.~(\ref{EqIterReWeightLeastSquare}) is quadratic in each iteration, it can be minimized by solving the set of linear equations:
\begin{equation}\label{EqLinearSystem}
\small{
\begin{split}
   &\left[(1-\alpha)\sum\limits_{j\in N_D(i)}\omega_{i,j}d_{i,j}^n+2\alpha\sum\limits_{j\in N_S(i)}\omega_{i,j}^{c}s_{i,j}^n\right]I_i\\
   &-2\alpha\sum\limits_{j\in N_S(i)}\omega_{i,j}^{g}s_{i,j}^nI_j=(1-\alpha)\sum\limits_{j\in N_D(i)}\omega_{i,j}d_{i,j}^nI_j^0
\end{split}
}
\end{equation}
we rewrite Eq.~(\ref{EqLinearSystem}) in matrix notation as:
\begin{equation}\label{EqLinearSystemMatrix}
\small{
\begin{split}
    &\ \ \ \ \ \ \ \ \ \left[(1-\alpha)W^n-2\alpha S^n\right]I = (1-\alpha)Z^nI^0\\
    &\Longrightarrow I^{n + 1}=(1-\alpha)\left[(1-\alpha)W^n-2\alpha S^n\right]^{-1}Z^nI^0
\end{split}
}
\end{equation}
where $I$ and $I^0$ are the vector form of the updated image and the input target image respectively. $W^n$ is a diagonal matrix with $W^n_{i,i}=\sum\limits_{j\in N_D(i)}\omega_{i,j}d^n_{i,j}+\frac{2\alpha}{1-\alpha}\sum\limits_{j\in N_S(i)}\omega^g_{i,j}s^n_{i,j}$. $S^n$ is the affinity matrix whose elements are $S^n_{i,j}=\omega^g_{i,j}s^n_{i,j}$. $Z^n$ is another affinity matrix with $Z^n_{i,j} = \omega_{i,j}d^n_{i,j}$.

\begin{figure*}
\centering
  \includegraphics[width=1\linewidth]{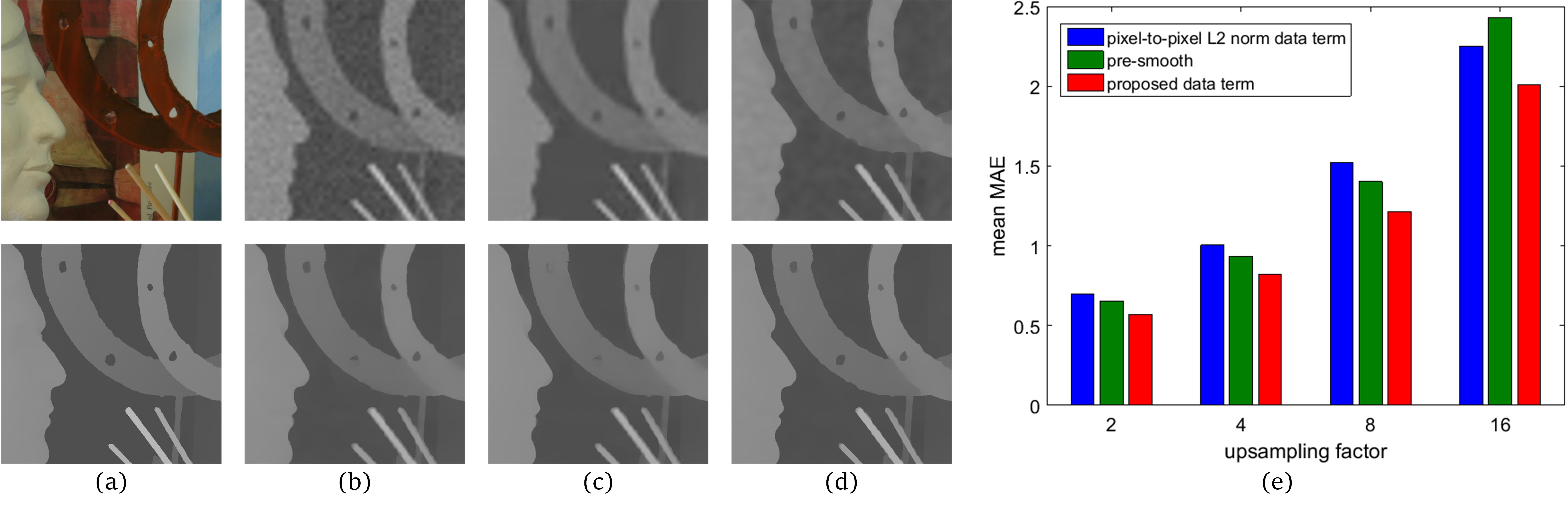}\\
  \caption{(a) Color image and groundtruth. (b) Bicubic interpolation of input depth map (in the first row) and the corresponding result (in the second row) using the method in \cite{liu2017robust} (i.e., pixel-to-pixel $L_2$ norm data term). (c) Pre-smooth the input depth map with BM3D \cite{dabov2007image}. The bicubic interpolation of the pre-smoothed depth map (in the first row) and the corresponding result (in the second row) using the method in \cite{liu2017robust}. (d) The input smoothed by the proposed data term (in the first row) and the corresponding result (in the second row) of our method. Example results are obtained on $8\times$ upsampling. (e) Quantitative comparison of different strategies on different upsampling factors in terms of MAE.}\label{FigDifferentDataTermCompDepthUpsampling}
\end{figure*}
\begin{figure}
\centering
  \includegraphics[width=1\linewidth]{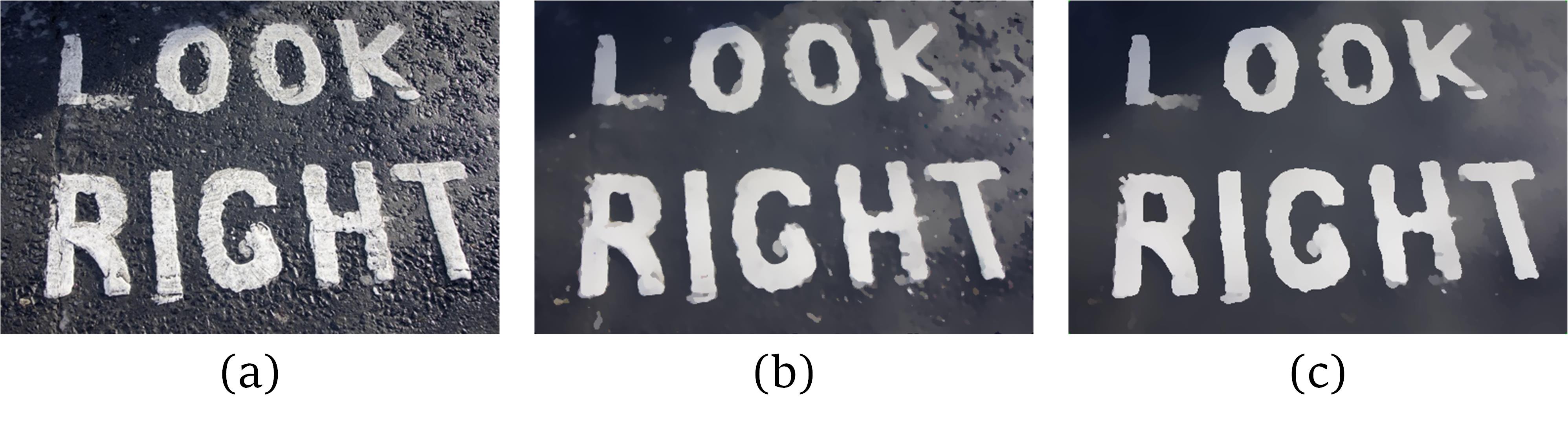}\\
  \caption{(a) Input image. Result of texture smoothing using (b) the method in \cite{liu2017robust} and (c) the proposed method.}\label{FigDifferentDataTermCompTextureSmooth}
\end{figure}

Solving the linear equation in Eq.~(\ref{EqLinearSystemMatrix}) has long been studied in the literature and there are many efficient modern solvers. In our experiment, we use the Preconditioned Conjugate Gradient (PCG) \cite{krishnan2013efficient} to solve Eq.~(\ref{EqLinearSystemMatrix}) which shows to be very efficient and can produce good results. Besides, our IRLS also has good convergence property. Fig.~\ref{FigCovergeAnalysis} shows the convergence analysis on guided depth map upsampling (dual-image task) and texture smoothing (single-image task) for different $\alpha$ values. Fig.~\ref{FigCovergeAnalysis}(a) shows Mean Absolute Errors (MAE) between the newly updated depth map and the groundtruth in each iteration on $8\times$ upsampling. Other upsampling factors have similar performance. Fig.~\ref{FigCovergeAnalysis}(b) shows Mean Absolute Difference (MAD) between two adjacent iterations in each iteration on texture smoothing. It is clear that our IRLS shows good convergence property for a large range of $\alpha$ values for different tasks. Note that the $\alpha$ value used in the experiments is based on the consideration of noise/texture smoothing, edges and small structures preserving. Small $\alpha$ cannot properly smooth the noise/texture while large $\alpha$ can over smooth edges and small structures. Fig.~\ref{FigCovergeAnalysis}(c) illustrates three example results of different $\alpha$ values on $8\times$ guided depth map upsampling and texture smoothing.

\subsection{Analysis}
\label{SecAnalysis}

In this section, we show further analysis on why our model are capable of handling the three practical issues discussed in Sec.~\ref{SecIntroduction}, namely, the heavy noise in input target images, texture copy artifacts and blurring edges caused by the structure inconsistency between the guidance image and the target image. Moreover, we show that the special structure of our model makes it capable of single-image tasks and promising results can be obtained. To make the analysis simple, we base our analysis on guided depth map upsampling for dual-image tasks and texture smoothing for single-image tasks. More application and the corresponding results will be shown in the experimental part in Sec.~\ref{SecAppResults}.

First, note that on the right side of Eq.~(\ref{EqLinearSystem}) is $\sum\limits_{j\in N_D(i)}\omega_{i,j}d_{i,j}^nI_j^0$ for each pixel $i\in \Omega$. This is in fact the filtered output of the input target image $I^0$ which results from our novel data term. If we use the data term that measures the $L_2$ norm of the pixel-to-pixel difference, then the right side of Eq.~(\ref{EqLinearSystem}) will be only $I^0_i$ for each pixel $i\in \Omega$. On the contrary, the filtering on the right side of Eq.~(\ref{EqLinearSystem}) can well smooth the noise in $I^0$. This makes our model more robust against the noise. In fact, the proposed data term is equivalent to the pixel-to-pixel $L_2$ norm data term of which the input image is a filtered image as the right side of Eq.~(\ref{EqLinearSystem}). In the field of guided depth map upsampling, there are approaches in the literature that independently denoised depth maps as a pre-processing step \cite{park2011high, xie2016edge}. Their data term still measures pixel-to-pixel difference with $L_2$ norm. However, different from these strategies that separately perform denoising and GIF, our data term can simultaneously perform denoising and GIF.

To further validate the effectiveness of the proposed data term, we perform another two experiments in guided depth map upsampling: (I) Upsample noisy input depth maps with the method in \cite{liu2017robust} that adopts the pixel-to-pixel $L_2$ norm data term. (II) First denoise input depth maps with BM3D \cite{dabov2007image} and then perform experiments with the denoised depth maps using the method in \cite{liu2017robust}. Fig.~\ref{FigDifferentDataTermCompDepthUpsampling}(b)$\sim$(d) show visual comparison on $8\times$ upsampling. The bicubic interpolated input depth map in Fig.~\ref{FigDifferentDataTermCompDepthUpsampling}(b) contains heavy noise and depth edges are also heavily blurred. The pre-smoothed depth map in Fig.~\ref{FigDifferentDataTermCompDepthUpsampling}(c) contains much less noise. However, small structures and depth edges are also heavily blurred. As a result, small structures on its upsampled depth map are completely blurred. Though the filtered input by our data term in Fig.~\ref{FigDifferentDataTermCompDepthUpsampling}(d) is a little more noisy than the one in Fig.~\ref{FigDifferentDataTermCompDepthUpsampling}(c), it contains less noise than the one in Fig.~\ref{FigDifferentDataTermCompDepthUpsampling}(b). Moreover, its depth edges are much sharper than the ones in both Fig.~\ref{FigDifferentDataTermCompDepthUpsampling}(b) and (c). This is because our filtering weights $d_{i,j}^n$ is based on the newly updated depth map. Small structures are also well preserved in our upsampled results. Fig.~\ref{FigDifferentDataTermCompDepthUpsampling}(e) further shows quantitative comparison on different upsampling factors. As shown in the figure, the proposed data term clearly outperforms compared methods which also validates our analysis.

In fact, the special structure of the data term can also benefit single-image tasks such as texture smoothing \cite{xu2012structure} due to its smoothing property described above. Fig.~\ref{FigDifferentDataTermCompTextureSmooth} (b) shows a result of texture smoothing using the method in \cite{liu2017robust} where texture in the image is not well smoothed. Fig.~\ref{FigDifferentDataTermCompTextureSmooth} (c) shows a result of texture smoothing with the proposed method where texture in the image is well smoothed and shows better visual quality than the one in Fig.~\ref{FigDifferentDataTermCompTextureSmooth} (b).

\begin{figure*}
\centering
  \includegraphics[width=1\linewidth]{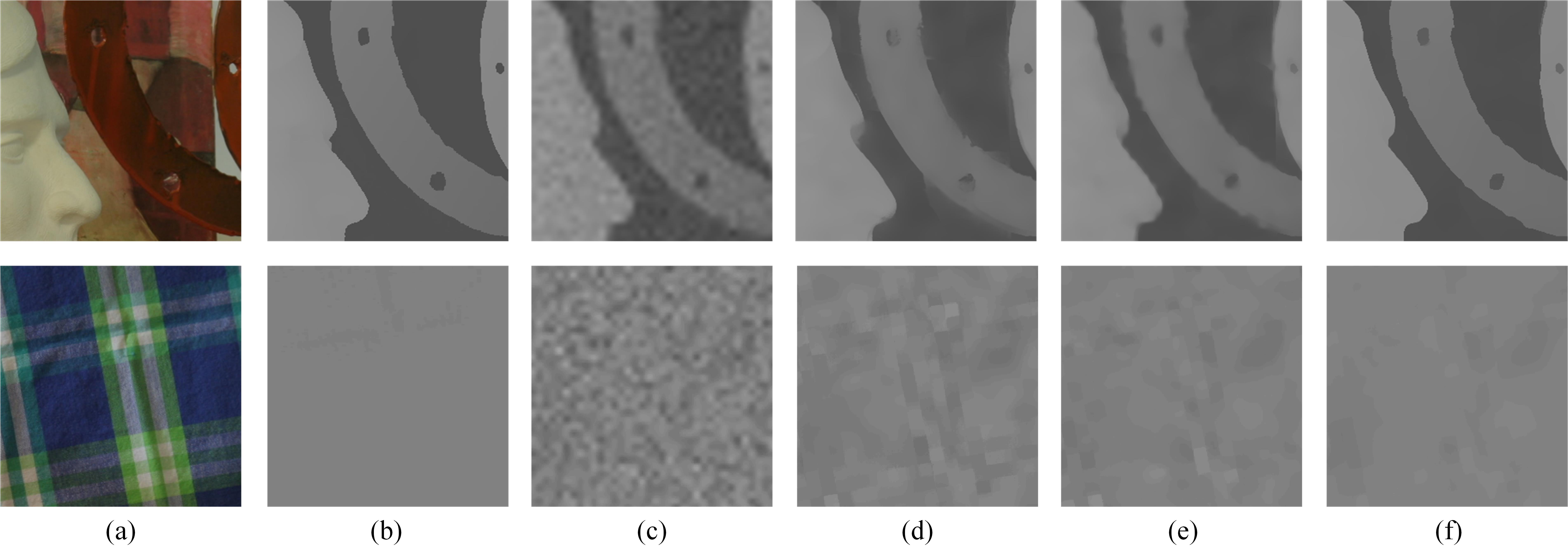}\\
  \caption{Comparison of depth edges preserving and texture copy artifacts suppression on $8\times$ upsampling. (a) Guidance color images. (b) Groundtruth. (c) Bicubic interpolation of input depth maps. Results obtained by the WLS \cite{min2014fast}, (e) the color guided AR model \cite{yang2014depth_recovery} and (f) our method. The first row shows depth edges preserving comparison. The second row shows texture copy artifacts suppression comparison.}\label{FigMotivation}
\end{figure*}

To handle the structure inconsistency between the guidance image and the target image, the key idea of our method is to make use of the property of not only the guidance image but also the target image. This idea is inspired by the work in guided depth map upsampling \cite{yang2014depth_recovery} which makes use of the bicubic interpolation of the input depth map to design the guidance weigh. However, our method can better preserve depth edges than theirs. This is because making use of bicubic interpolated input depth maps can help to suppress texture copy artifacts but may fail to properly preserve sharp depth edges when the upsampling factor is large and the input depth map contains heavy noise. In this case, depth edges have already been blurred as illustrated in Fig.~\ref{FigMotivation}(c). In contrast, the guide weight of the smoothness term of our IRLS in Eq.~(\ref{EqIterReWeightLeastSquare}) $s_{i,j}^n$ is based on the newly updated depth map. Its quality is much better than the bicubic interpolation with much less noise and sharper depth edges. In this way, the final weight of the smoothness term in our IRLS in Eq.~(\ref{EqIterReWeightLeastSquare}), i.e., $\omega_{i,j}^gs_{i,j}^n$, can make better use of the property of the depth map as well as the guidance image. Thus, it can not only suppress texture copy artifacts but also better preserve sharp depth edges. Fig.~\ref{FigMotivation} shows the comparison of three different methods: 1) the Weighted Least Squares (WLS) \cite{min2014fast} whose guidance weight is only based on the guidance image\footnote{We use the same formulation as defined by \cite{min2014fast} but with a larger $2r_s+1\times2r_s+1$ neighborhood which is the same as ours. We use the same bilateral guidance weight as defined in Eq.~(\ref{EqColorSpatialWeight}) for its smoothness term.}, 2) the color guided Auto-Regressive (AR) model \cite{yang2014depth_recovery} whose guidance weight is based on both the guidance image and the bicubic interpolation of the input depth map, 3) our method whose guidance weight is based on the guidance image and the newly updated depth map. The WLS \cite{min2014fast} produces results with blurred depth edges and heavy texture copy artifacts. The color guided AR model \cite{yang2014depth_recovery} can properly preserve depth edges but they are still not sharp. There are still texture copy artifacts in the results. Our method can preserve sharp depth edges and there are seldom texture copy artifacts in our results.

\begin{algorithm}
\caption {Robust Guided Image Filtering with Parameter Optimization}\label{Alg}
\begin{algorithmic}[1]
\REQUIRE ~~\\
$I^0$: input image\\

$G$: guidance image\\

$\alpha$: parameter in Eq.~(\ref{EqOurModel})

$r_D$,$r_S$: radius of neighborhood for the data term and the smoothness term\\

$\sigma_d$,$\sigma_s$,$\sigma_g$: parameters for $\omega_{i,j}$ in Eq.~(\ref{EqSpatialWeight}) and $\omega_{i,j}^g$ in Eq.~(\ref{EqColorSpatialWeight})\\

$\beta$: parameter in Eq.~(\ref{EqParameterModel})\\

$\lambda^0$: initial parameter for Eq.~(\ref{EqParameterUpdate})\\

$\tau$: updating rate in Eq.~(\ref{EqParameterUpdate})\\

\ENSURE ~~\\
$I^\ast$: filtered image
\WHILE{Eq.~(\ref{EqLinearSystemMatrix}) does not converge to the fixed point}
\STATE Update image using Eq.~(\ref{EqLinearSystemMatrix})
\STATE Update parameter using Eq.~(\ref{EqParameterUpdate})
\ENDWHILE
\RETURN $I^\ast$
\end{algorithmic}
\end{algorithm}
\subsection{Data Driven Parameter Optimization}
\label{SecParameterSelection}

In all the applications in this paper, guided depth map upsampling is among the most challenging tasks. This is because despite the challenging issues mentioned in Sec.~\ref{SecIntroduction}, the resolution of input depth maps is smaller than guidance color images. As the upsampling factor becomes large, how to preserve depth edges that correspond to weak color edges and small structures also becomes challenging. We propose a data-driven parameter optimization framework to properly handle this problem. In fact, the parameter $\lambda$ in Eq.~(\ref{EqErroNormFunction}) is an important parameter in our model. A large $\lambda $ can better smooth noise but may blur depth edges. Thus, pixels in smooth regions on the depth map should be assigned with large $\lambda$. A small $\lambda $ can better preserve depth edges but performs poorly in noise smoothing. Thus, pixels along depth edges should be assigned with small $\lambda$. To eliminate heuristic parameter selection, we describe another data driven adaptive parameter optimization model that adapts $\lambda $ to each pixel on the depth map. Because the depth map is piece-wise smooth, we assume that $\lambda$ is also regular and smooth. Therefore, we add another term that consists of the $L_2$ norm of the gradient of ${\lambda}_{i}(i\in\Omega)$ to the objective function in Eq.~(\ref{EqOurModel}), resulting in the following objective function:
\begin{equation}\label{EqParameterModel}
\small
{
\begin{split}
    E(\lambda)=(1-\alpha)E_D(I,I^0)+\alpha E_S(I)+\beta \sum\limits_{i\in \Omega }{|\nabla\lambda_i|^2},\ i\in \Omega
\end{split}
}
\end{equation}
By minimizing Eq.~(\ref{EqParameterModel}) with respect to ${{\lambda }_{i}}$ through the steepest gradient descent according to:
\begin{equation}\label{EqParameterUpdate}
\small{
    \lambda _{i}^{n+1}=\lambda _{i}^{n}-\tau \frac{\partial E}{\partial \lambda _{i}^{n}},\ i\in \Omega
    }
\end{equation}
where $\tau $ is a given updating rate and the derivative of the objective function is given by
\begin{equation}\label{EqParameterDerivative}
\small
{
\begin{split}
    &\frac{\partial E}{\partial \lambda^n_i} = \\
    &( 1-\alpha)\sum\limits_{j\in N_D(i)}{{{\omega }_{i,j}}\left[ 4{{\lambda }_{i}^n}\left( 1-{{d}_{i,j}^n} \right)-\frac{2{{\left( {{I}^n_{i}}-I_{j}^{0} \right)}^{2}}}{{{\lambda}_{i}^n}}{{d}^n_{i,j}} \right]}\\
    &+\alpha \sum\limits_{j\in N_S(i)}{{{{{\omega }}}^g_{i,j}}\left[ 4{{\lambda}_{i}^n}\left( 1-{{s}^n_{i,j}} \right)-\frac{2{{\left( {{I}^n_{i}}-{{I}^n_{j}} \right)}^{2}}}{{{\lambda }_{i}^n}}{{s}^n_{i,j}} \right]}\\
    &+2\beta \sum\limits_{i\in \Omega }{\Delta {{\lambda }_{i}^n}}
\end{split}
}
\end{equation}
where ${{d}_{i,j}^n}$ and ${{s}_{i,j}^n}$ are the same as Eq.~(\ref{EqErrorNormFunctionDerivative}).

\begin{figure}
\centering
  \includegraphics[width=1\linewidth]{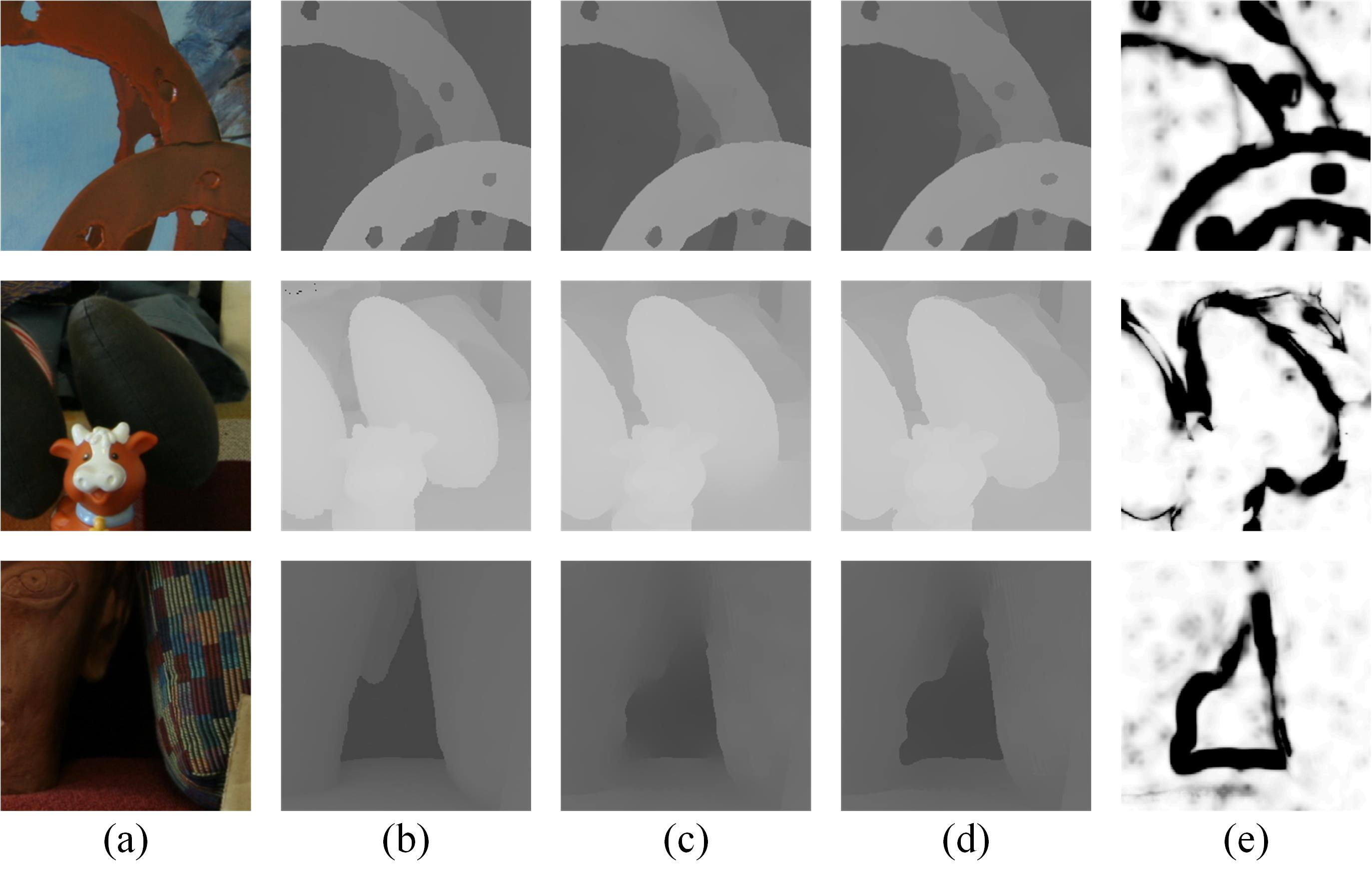}\\
  \caption{Visual comparison of our method for $8\times$ guided depth map upsampling with and without parameter optimization. (a) Guidance color images. (b) Groundtruth. (c) Results obtained without parameter optimization. (d) Results obtained with parameter optimization and (e) the corresponding parameter maps.}\label{FigParaSelectAdvantage}
\end{figure}
\begin{figure*}
\centering
  \includegraphics[width=1\linewidth]{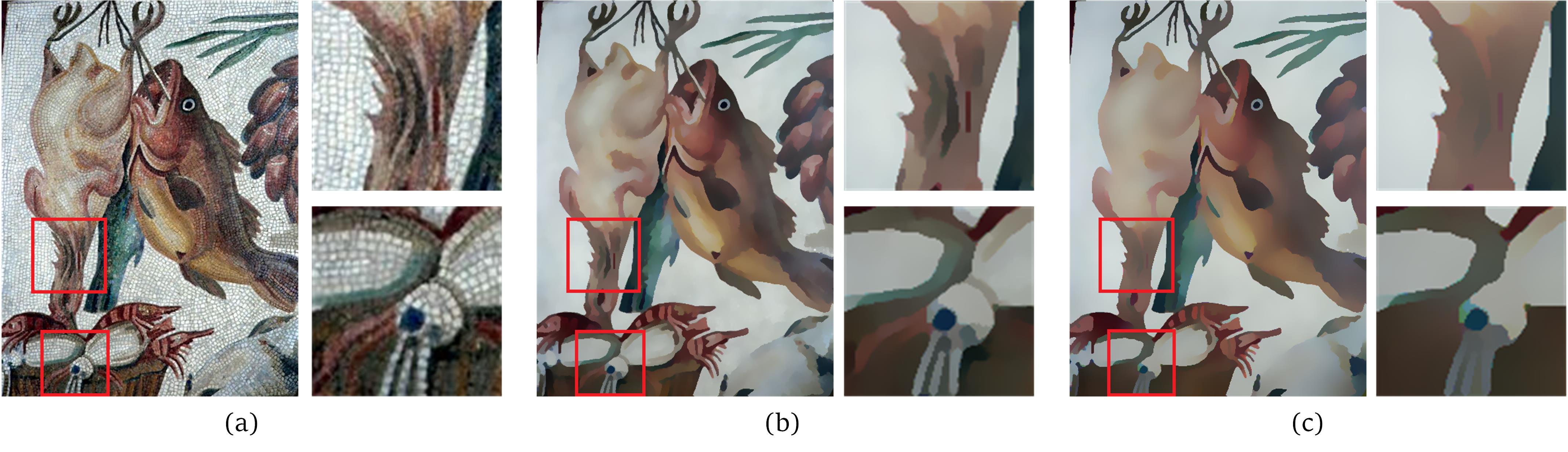}\\
  \caption{Visual comparison of our method for texture smoothing with and without guidance weight. (a) Input image. (b) Result of texture smoothing with guidance weight. (c) Result of texture smoothing without guidance weight. Regions in red boxes are highlighted.}\label{FigGuideWeightAdvantage}
\end{figure*}

In our experiments, depth map updating and the parameter optimization are addressed in an iterative fashion through alternating the parameter update in Eq.~(\ref{EqParameterUpdate}) and the depth map update in Eq.~(\ref{EqLinearSystemMatrix}). We summarize our method with parameter adaptation in algorithm \ref{Alg}. Fig.~\ref{FigParaSelectAdvantage}(e) illustrates several parameter maps obtained by our method. It is clear that the parameter maps well correspond to the character of depth maps shown in Fig.~\ref{FigParaSelectAdvantage}(b). The parameter adaptation around depth edges corresponds to lower values than smooth regions. Visual comparison illustrated in Fig.~\ref{FigParaSelectAdvantage}(c) and (d) shows that the proposed parameter optimization helps to preserve sharp depth edges and small structures even for $8\times$ upsampling.

\subsection{Relation to Other Methods}
\label{SecRelation2Others}
In this section, we show that several existing models are special cases of our model which can be obtained by varying parameters in our model. First of all, we should know that the exponential error norm in Eq.~(\ref{EqErroNormFunction}) and Eq.~(\ref{EqSoomthnessNormFun}) used in the data term and the smoothness term is not a new one proposed by us. It is known as Perona and Malik error norm function \cite{perona1990scale} which was originally used for robust image smoothing. It also has many variants \cite{mrazek2006robust, pizarro2010generalised}. However, most of the related work mainly focused on single-image tasks. As far as we know, we are the first to apply it to dual-image tasks, i.e., guided image filtering. We show that it yields promising results in many applications in handling the structure inconsistency between the guidance images and the target image. Moreover, as shown in this section, we are the first to show the relation between the adopted error norm function and the $L_2$ error norm which was not explored in the literature.

Among all the related methods, the one that is most close to our work is the Bayesian framework proposed by Mrazek et al. \cite{mrazek2006robust}. Their model can be obtained by replacing the guidance weight in Eq.~(\ref{EqColorSpatialWeight}) in our model with a Gaussian window. The main difference between our model and this Bayesian framework is that our model is applicable to both dual-image tasks and single-image tasks while this Bayesian framework is only suitable for single-image tasks. However, we show that even for single-image tasks such as texture smoothing, our method can better preserve edges than theirs. We show illustration in Fig.~\ref{FigGuideWeightAdvantage}. In addition, the proposed IRLS numeric solution for such kind of problem in Sec.~\ref{SecNumSolution} is also seldom proposed before which is also the contribution of this paper.

Recently, a new filter named Guided Bilateral Filter (GBF) was proposed by Caraffa et al. \cite{caraffa2015guided}. It has been show that GBF performs better in preserving edges than BF proposed by Tomasi et al.\cite{tomasi1998bilateral}. GBF can be obtained from our model by simply setting $\alpha=1$.

\begin{table*}
\centering
\caption{Parameter setting used in all the applications.}\label{TabParameter}
\resizebox{1\linewidth}{!}
{
\begin{tabular}{c|c|c|c|c|c|c|c|c|c|c}
  \Xhline{1.5pt}
  Application$\backslash$Parameters & $\alpha$ & $r_d$ & $r_s$ & $\sigma_d$ & $\sigma_s$ & $\sigma_g$ & $\lambda_d$ & $\lambda_s$ & $\beta$ & $\tau$ \\
  \Xhline{1.5pt}

  Guided Depth Updampling & 0.6($2\times$)/0.8($4\times$)/0.9($8\times$)/0.93($16\times$) & 7 & 7 & 7 & 7 & 10 & 7 & 7 & 0.5 & 0.3 \\
  \hline

  Flash/No Flash Filtering & 0.7 & 1 & 4 & 1 & 4 & 5 & 5 & 5 & - & - \\
  \hline

  Detail Enhancement & 0.8 & 0 & 3 & - & 3 & 10 & $\infty$ & 10 & - & -\\
  \hline

  HDR Tone Mapping & $\frac{1}{9}/\frac{1}{2}/\frac{8}{9}$(for the first, second and third detail layer) & 1 & 6 & 1 & 6 & 20 & 20 & 20 & - & -\\
  \hline

  Texture Smoothing & 0.9 & 5 & 5 & 5 & 5 & 15 & 10 & 10 & - & -\\
  \hline

  Clip-art Compression Artifacts Removal & 0.9 & 5 & 5 & 5 & 5 & 15 & 10 & 10 & - & - \\
  \Xhline{1.5pt}
\end{tabular}
}
\end{table*}

Our model is also closely related to the WLS model. The WLS is a basic optimization framework that was originally used for image deblurring \cite{lagendijk1988regularized}. Recently, it has also been used in multi-scale tone mapping and detail manipulation \cite{farbman2008edge}, stereo matching \cite{min2008cost} and guided depth map upsampling \cite{min2014fast}. The WLS is formulated as:
\begin{equation}\label{EqWLS}
\scriptsize
\begin{split}
    I^\ast=\underset{I}{\mathop{\arg}}\min\{(1-\alpha)\sum\limits_{i\in \Omega}{(I_i - I_j^0)^{2}}+\alpha \sum\limits_{i\in \Omega }{\sum\limits_{j\in N_S(i)}{\kappa _{i,j}{({I}_{i}-{I}_{j})^2}}}\}
\end{split}
\end{equation}

The WLS can also contain the aggregated data term instead of the pixel-to-pixel data term. The WLS with aggregated data term is defined as:
\begin{equation}\label{EqRWLS}
\small{
\begin{split}
    &I^\ast=\underset{I}{\mathop{\arg}}\min\{(1-\alpha)\sum\limits_{i\in \Omega}\sum\limits_{j\in N_D(i)}{c_{i,j}(I_i-I_j^0)^2}\\
    &\ \ \ \ \ \ \ \ \ \ \ \ \ \ \ \ \ \ \ \ \ \ \ \ \ \ \ \ \ \ \ \ \ \ +\alpha \sum\limits_{i\in \Omega }{\sum\limits_{j\in N_S(i)}{\kappa _{i,j}{({I}_{i}-{I}_{j})^2}}}\}
\end{split}
}
\end{equation}

In this section, we show that by slight varying the parameters in our model, we can rightly obtain the WLS in Eq.~(\ref{EqWLS}) and Eq.~(\ref{EqRWLS}). First, we show that by setting $\lambda\to\infty$, our exponential error norm in Eq.~(\ref{EqErroNormFunction}) and Eq.~(\ref{EqSoomthnessNormFun}) degraded to the $L_2$ error norm for any bounded input. We denote $\varphi(x^2, \lambda)=\varphi_D(x^2, \lambda) = \varphi_S(x^2, \lambda)$, $f(\lambda) = 1-\exp \left(-\frac{x^2}{2\lambda^2}\right)$ and $g(\lambda) = \frac{1}{2\lambda^2}$. According to L'Hopital's rule, we have:
\begin{eqnarray}\label{EqTaylorSeries}
\small{
\begin{split}
&\lim_{\lambda\to\infty}\varphi(x^2, \lambda) = \lim_{\lambda\to\infty}\frac{f(\lambda)}{g(\lambda)} =\lim_{\lambda\to\infty}\frac{\frac{\partial{f(\lambda)}}{\partial{\lambda}}}{\frac{\partial{g(\lambda)}}{\partial{\lambda}}}\\
& \ \ \ \ \ \ \ \ \ \ \ \ \ \ \ \ \ \ \ \ \ = \lim_{\lambda\to\infty}\frac{-\exp(-\frac{x^2}{2\lambda^2})\cdot\frac{x^2}{\lambda^3}}{-\frac{1}{\lambda^3}}\\
& \ \ \ \ \ \ \ \ \ \ \ \ \ \ \ \ \ \ \ \ \ = \lim_{\lambda\to\infty}\exp(-\frac{x^2}{2\lambda^2})\cdot x^2\\
& \ \ \ \ \ \ \ \ \ \ \ \ \ \ \ \ \ \ \ \ \ = x^2
\end{split}
}
\end{eqnarray}

Based on this, we show that the WLS in Eq.~(\ref{EqWLS}) and Eq.~(\ref{EqRWLS}) are special cases of our model in Eq.~(\ref{EqOurModel}) by varying its $r_d$ and $\lambda$:
\begin{itemize}
  \item[1.] By setting $r_d=0$ and $\lambda\to\infty$, we obtain the WLS in Eq.(\ref{EqWLS}) with $\kappa_{i,j}=\omega_{i,j}^c$.

  \item[2.] By setting $\lambda\to\infty$, we obtain the WLS in Eq.(\ref{EqRWLS}) with $\kappa_{i,j}=\omega_{i,j}^c$ and $c_{i,j}=\omega_{i,j}$.
\end{itemize}

The above two cases correspond to the WLS that only uses the guidance color image for the guidance weight. In guided depth map upsampling, there are also methods \cite{yang2014depth_recovery} that combine the guidance color image and the bicubic interpolation of the input depth map to design the guidance weight. We show that this can also be obtained from our model. To make the analysis more clear, we denote $\lambda$ in the data term and the smoothness term as $\lambda_d$ and $\lambda_s$ respectively. If we use the bicubic interpolation of input depth maps as initialization, we have:

\begin{itemize}
  \item[1.] By setting $r_d=0$ and $\lambda_d\to\infty$, the first iteration of our IRLS in Eq.~(\ref{EqIterReWeightLeastSquare}) is the WLS in Eq.(\ref{EqWLS}) with $\kappa_{i,j}=\omega_{i,j}^cs_{i,j}^0$.

  \item[2.] By setting $\lambda_d\to\infty$, the first iteration of our IRLS in Eq.~(\ref{EqIterReWeightLeastSquare}) is the WLS in Eq.(\ref{EqRWLS}) with $\kappa_{i,j}=\omega_{i,j}^cs_{i,j}^0$ and $c_{i,j}=\omega_{i,j}$.
\end{itemize}

All the above analyses together cast new insights into the advantages of the proposed method. They also help to have a better understanding on how the proposed method outperforms previous methods in handling the challenging issues in GIF.

\section{Applications and Experimental Results}
\label{SecAppResults}

Our method avails several applications including not only dual-image tasks but also single-image tasks due to its special properties in image filtering. We apply it to guided depth map upsampling, flash/no flash filtering, detail enhancement, multi-scale HDR tone mapping, texture smoothing and clip-art JPEG compression artifacts removal. Parameter adaptation described in Sec.~\ref{SecParameterSelection} is only applied to guided depth map upsampling. All the images are firstly normalized into $[0, 255]$ before they are filtered and then normalized back into their original range. Parameter setting in all the applications are shown in Tab.~\ref{TabParameter}.

\begin{figure*}
\centering
  \includegraphics[width=1\linewidth]{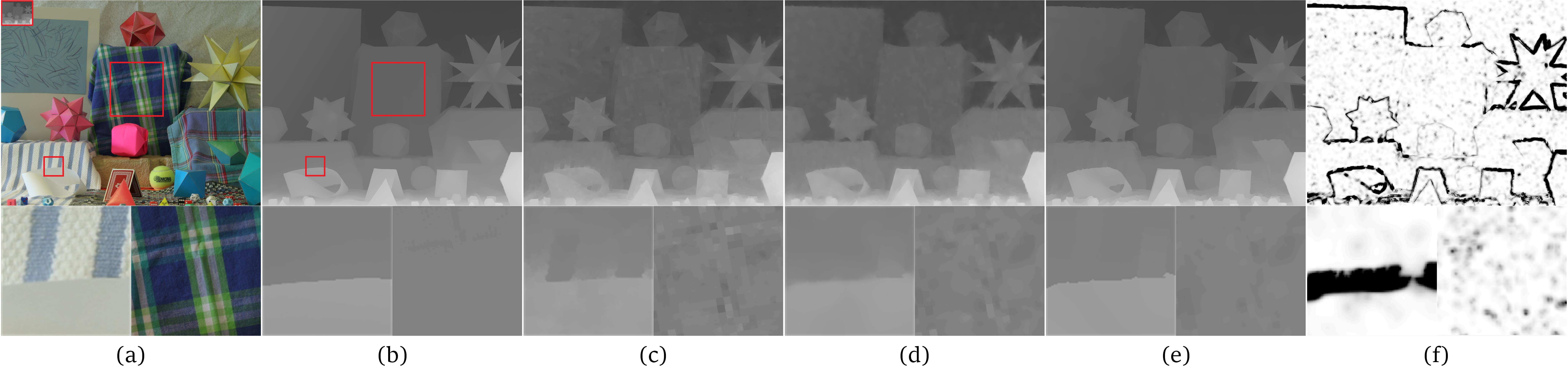}\\
  \caption{$8\times$ guided depth map upsampling results. (a) Input depth map (in red box) and the corresponding guidance color image. (b) Groundtruth depth map. Result of (c) NLM-WLS \cite{park2011high}, (d) color guided AR model \cite{yang2014depth_recovery} and (e) our method. (f) The corresponding parameter map by our parameter optimization. Regions in red boxes are highlighted.}\label{FigToFSimulated}
\end{figure*}
\begin{figure*}
\centering
  \includegraphics[width=1\linewidth]{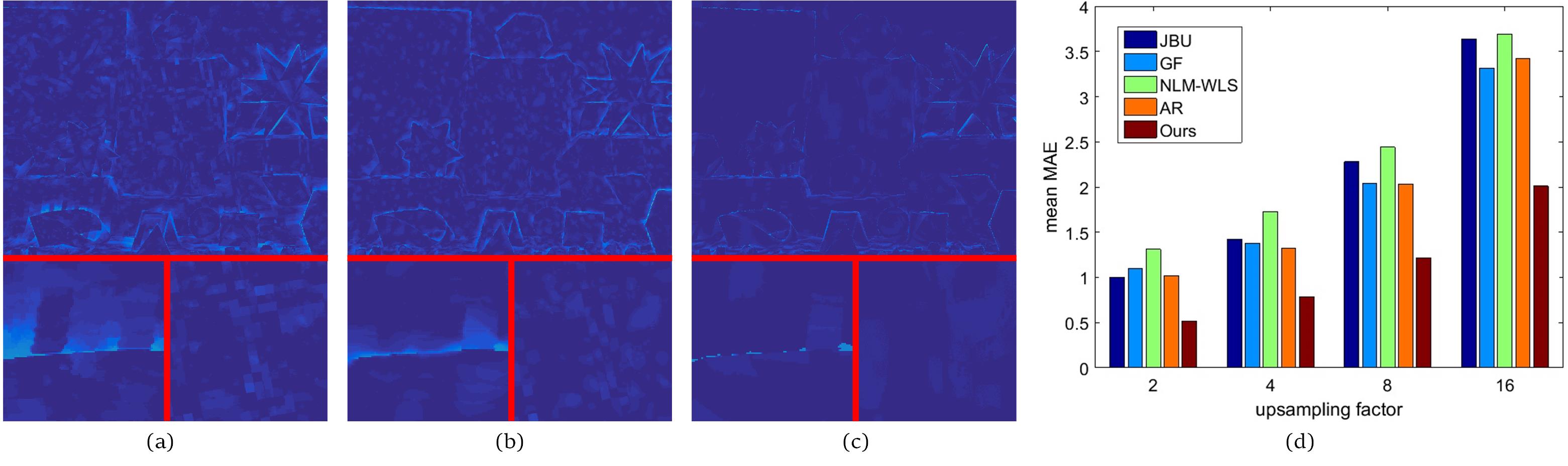}\\
  \caption{Error maps of the results in Fig.~\ref{FigToFSimulated}. (a) NLM-WLS \cite{park2011high}, (b) color guided AR model \cite{yang2014depth_recovery}, (c) our method, (d) mean MAE of different methods.}\label{FigToFSimulatedDiff}
\end{figure*}

\subsection{Guided Depth Map Upsampling}
\label{SecGuidedDepthUpsampling}

Depth maps captured by modern ToF depth cameras are usually contaminated by heavy noise and suffer from being of low resolution. Guided depth map upsamping aims to generate a high-resolution noise free depth map by using a high-resolution color image and a noisy low resolution depth map. Upsampled depth maps should avoid texture copy artifacts and preserve sharp depth edges. The heavy noise should also be well smoothed. The above goals become increasingly challenging as the updampling factor becomes large (e.g., $8\times$).

To handle the structure inconsistency, most state-of-the-art methods adopt two strategies: (I) using complex guidance weight and (II) making use of the bicubic interpolation of the noisy low-resolution depth map. Park et al.\cite{park2011high} proposed to incorporate different weighting schemes to combine different cues including segmentation, image gradients, edge saliency and the bicubic interpolation of input depth maps. The auto-regressive coefficient of the color guided Auto-Regressive (AR) model \cite{yang2014depth_recovery} consists of the combination of the "shape-based" color guidance weight and the weight based on the bicubic interpolated input depth map. However, more complex guidance weight means heavier computational cost. More importantly, it can also result in heavier texture copy artifacts.  Also, the bicubic interpolation of the input depth map becomes unreliable especially when the upsampling factor is large (e.g., $8\times$) and the input depth map contains heavy noise.

On the contrary, we only adopt the simple bilateral guidance weight defined in Eq.~(\ref{EqColorSpatialWeight}). We show that simple bilateral guidance weight is enough to achieve sharp depth edges while avoiding texture copy artifacts due to the special structure of our smoothness term. We have shown detailed analysis in Sec.~\ref{SecAnalysis}. Visual comparison is shown in Fig.~\ref{FigToFSimulated} with examples of $8\times$ upsampling results. We use the bicubic interpolation of the input depth map as the initial input of our model. Our method is compared with the weighted least squares proposed by Park et al.\cite{park2011high} that we denote as NLM-WLS, the color guided AR model proposed by Yang et al. \cite{yang2014depth_recovery}\footnote{We re-implement the AR model with C language for time and memory efficiency based on the author provided MATLAB source code. The author provided source code is available here: {\scriptsize\url{http://cs.tju.edu.cn/faculty/likun/projects/depth_recovery/index.htm}}. Our implementation is available here: {\scriptsize\url{https://github.com/wliusjtu/Adaptive-Auto-regressive-Model-for-Guided-Depth-Map-Restoration}}.}. The corresponding Mean Absolute Error (MAE) maps are illustrated in Fig.~\ref{FigToFSimulatedDiff}. It is clear our method can better smooth the noise. Quantitative measurement of mean MAE of each upsampling factor in Fig.~\ref{FigToFSimulatedDiff}(d) also validates the effectiveness of our method.

\begin{figure*}
\centering
  \includegraphics[width=1\linewidth]{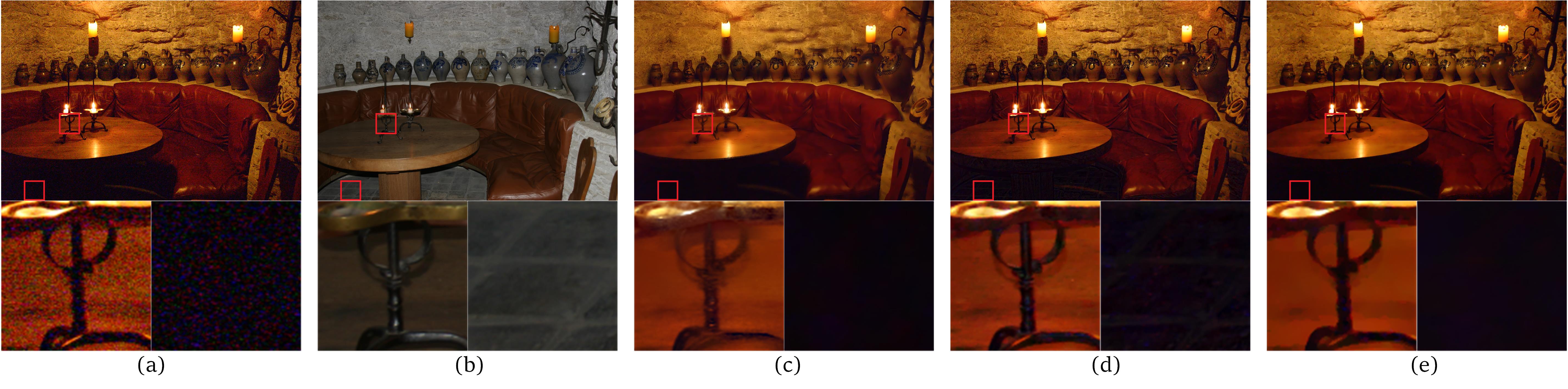}\\
  \caption{Flash/no flash filtering results. (a) No flash image. (b) Flash image. Result of (c) GF \cite{he2013guided}, (d) Shen et al. \cite{shen2015multispectral} and (e) our method. Regions in red boxes are highlighted.}\label{FigFlashNoFlash}
\end{figure*}

\subsection{Flash/No Flash Filtering}
\label{SecFlashNoFlash}

\begin{figure*}
\centering
  \includegraphics[width=1\linewidth]{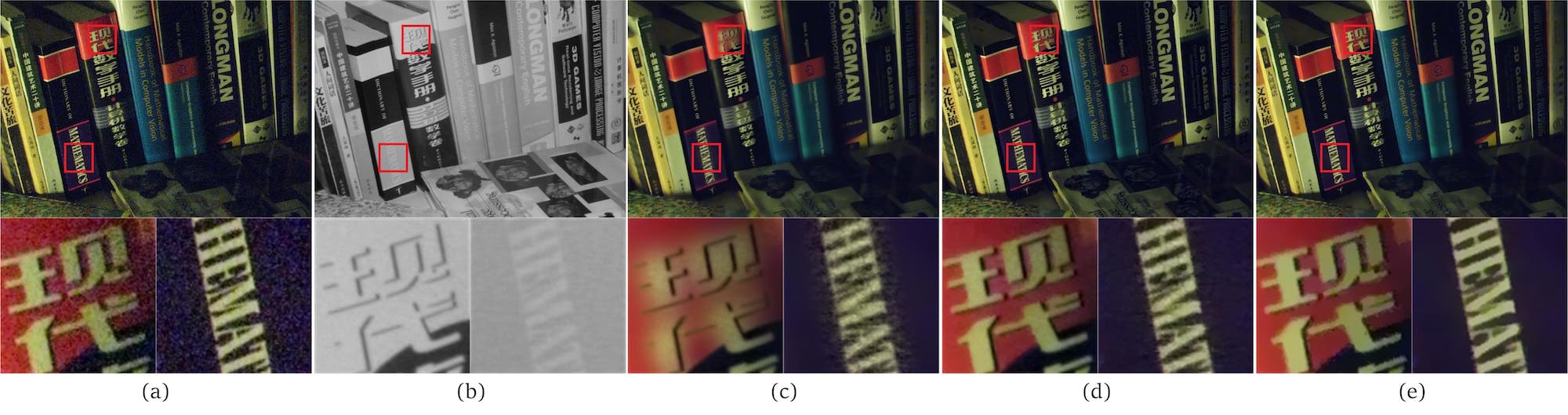}\\
  \caption{NIR image guided color image filtering results. (a) Color image. (b) NIR image. Result of (c) GF \cite{he2013guided}, (d) Shen et al. \cite{shen2015multispectral} and (e) our method. Regions in red boxes are highlighted.}\label{FigNIRGuiedImageFiltering}
\end{figure*}

Image captured under low light condition can contain noise but it can capture the ambient illumination. Instead, one can capture another image with flash light at the same place which is more noiseless and contains more details and sharper edges. In this case, the flash image can be used as guidance to enhance the quality of the no flash image \cite{petschnigg2004digital, eisemann2004flash}. Work proposed by Krishnan et al. \cite{krishnan2009dark} used dark flash to capture NIR images to guide the denoising process. The main challenge is shadows in the flash image caused by flash that do not exist in the no flash image, i.e., structure inconsistency. This may result in texture copy artifacts or blurring edges in results. We show such an example in Fig.~\ref{FigFlashNoFlash}. However, we find for methods with strong texture copy ability such as the one proposed by Shen et al. \cite{shen2015multispectral}, structures of objects that appear in flash images due to flash but disappear in no flash images can also cause texture copy artifacts in filtered images. We show such an example of the "floor" in the image in the highlighted region in Fig.~\ref{FigFlashNoFlash}(d). Despite the shadow problem, edges in flash NIR images can become quite weak or even disappear. This is illustrated in Fig.~\ref{FigNIRGuiedImageFiltering}. In this case, blurring edges will occur in the result as illustrated in Fig.~\ref{FigNIRGuiedImageFiltering} (c) and (d).

\begin{figure*}
\centering
  \includegraphics[width=1\linewidth]{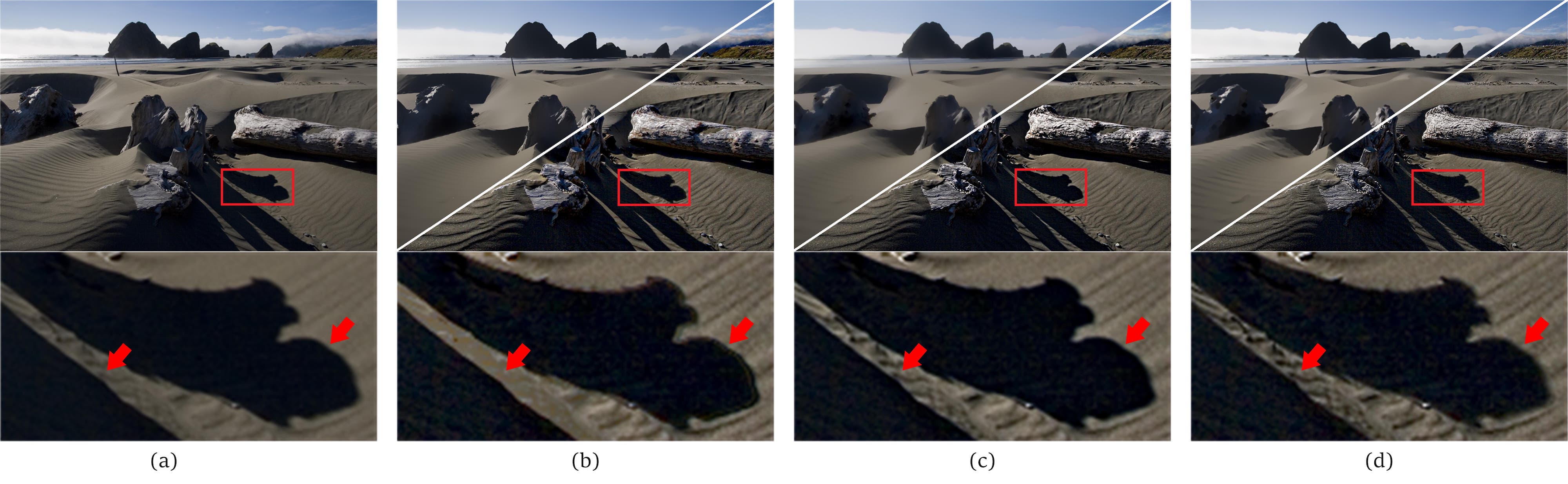}\\
  \caption{Detail enhancement results. (a) Input image. Result of (b) adaptive manifold filtering \cite{gastal2012adaptive}, (c) WLS \cite{farbman2008edge} and (d) our method. The left up part in each image of (b)$\sim$ (d) is the smoothed base layer. The right down part in each image is the enhanced image. Regions in red boxes are highlighted.}\label{FigDetailEnhance}
\end{figure*}
\begin{figure*}
\centering
  \includegraphics[width=1\linewidth]{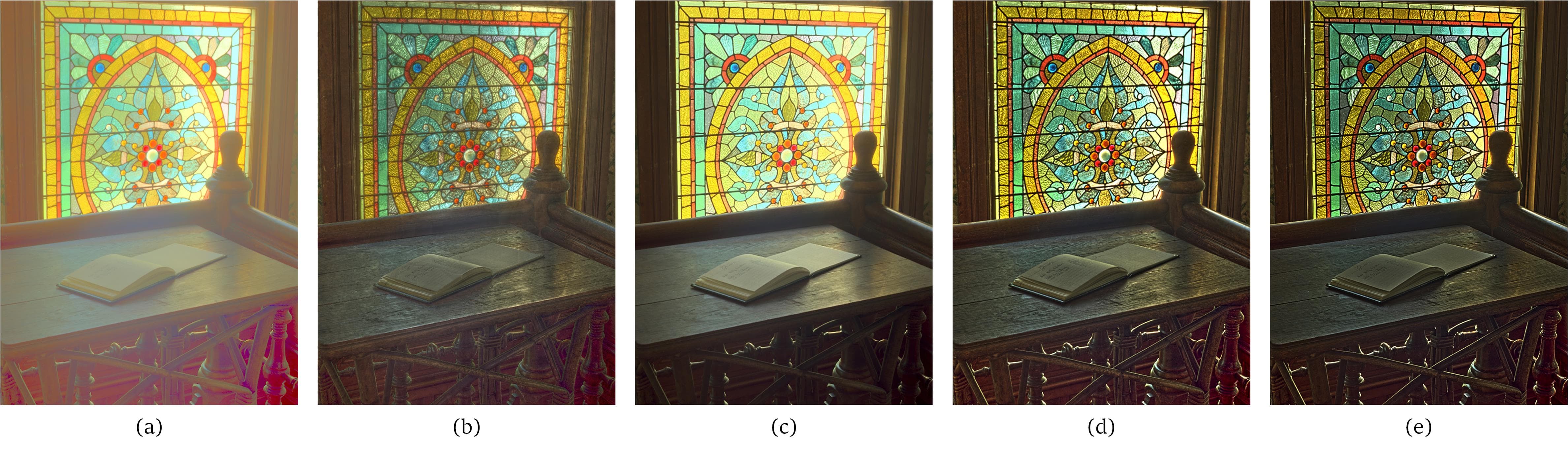}\\
  \caption{HDR tone mapping Results. (a) Original HDR image. Result of (b) Xu et al.\cite{xu2011image}, (c) Reinhard et al. \cite{reinhard2002photographic}, (d) WLS \cite{farbman2008edge}, (e) Ours.}\label{FigHDRToneMapping}
\end{figure*}

For the shadow problem, various shadow detection methods were proposed to eliminate such structure inconsistency \cite{petschnigg2004digital, eisemann2004flash, krishnan2009dark}. However, for the rest problems mentioned above, there are few related methods proposed to handle them. The key problem of previous method \cite{petschnigg2004digital, eisemann2004flash, krishnan2009dark, shen2015multispectral} is that their methods heavily rely the structure information on the guidance image. On the contrary, the idea of our method is to make use of the property of both the guidance image and the target image. This is very useful to handle the problems above. We show our results in Fig.~\ref{FigFlashNoFlash}(e) and Fig.~\ref{FigNIRGuiedImageFiltering}(e) where the above problems are well handled.

\subsection{Layer Decomposition Based Manipulation}

In applications such as detail enhancement and HDR tone mapping, images are decomposed into a base layer and a detail layer (or more detail layers). Then images are enhanced by firstly manipulating either the base layer or the detail layer and then re-combining them. The base layer is usually obtained by smoothing the original image with an edge-preserving filter such as BF \cite{fattal2007multiscale, durand2002fast}, WLS \cite{farbman2008edge} and gradient $L_0$ norm optimization \cite{xu2011image}. It is required to preserve edges to avoid halos and gradient reversals \cite{farbman2008edge, he2013guided}. Usually, global methods \cite{farbman2008edge, xu2011image} are superior to local methods \cite{fattal2007multiscale, durand2002fast} in handling these problems. Fig.~\ref{FigDetailEnhance}(b) shows an example of gradient reversals in image detail enhancement produced by adaptive manifold filtering \cite{gastal2012adaptive} which is a local method. In addition, textures labeled with the red arrow also disappear in the result of adaptive manifold filtering \cite{gastal2012adaptive} in Fig.~\ref{FigDetailEnhance}(b) which is due to insufficient smoothing. On the contrary, WLS \cite{farbman2008edge} and our method can avoid such problems with details properly enhanced.

HDR tone mapping is also achieved by layer decomposition \cite{li2005compressing, durand2002fast, farbman2008edge}. The base layer is nonlinearly mapped to a low dynamic range and is re-combined with the detail layer. Under the multi-scale HDR tone mapping framework proposed by Farbman et al. \cite{farbman2008edge}\footnote{The source code can be downloaded here \url{http://www.cs.huji.ac.il/~danix/epd/}}, we use our model for layer decomposition, which is applied to the logarithmic HDR images. Results are illustrated in Fig.~\ref{FigHDRToneMapping} (e) where structures are well preserved or enhanced. Results of other methods are also illustrated in Fig.~\ref{FigHDRToneMapping}(b)$\sim$(d).

\begin{figure*}
\centering
  \includegraphics[width=1\linewidth]{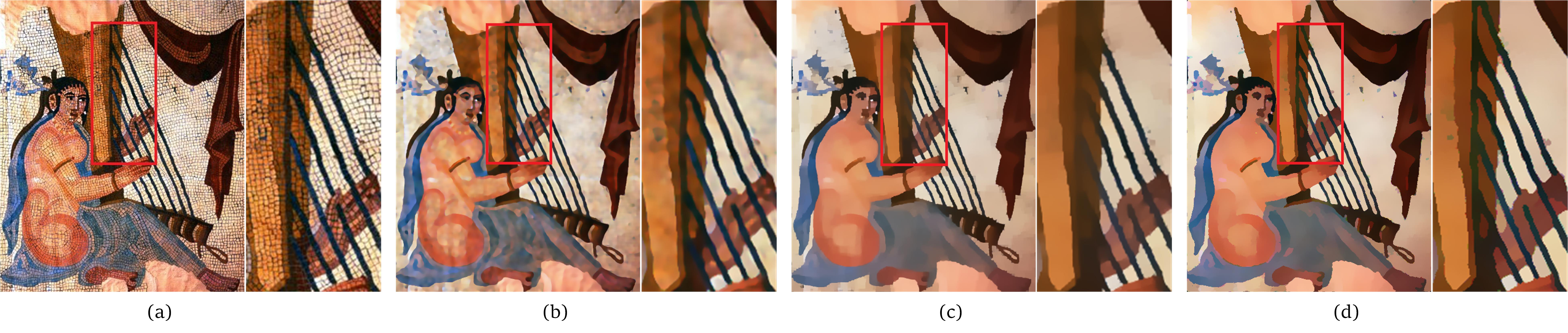}\\
  \caption{Texture smoothing results. (a) Input image. Result of (b) RGF \cite{zhang2014rolling}, (c) RTV \cite{xu2012structure} and (d) our method. Regions in red boxes are highlighted.}\label{FigTextureSmooth}
\end{figure*}
\begin{figure*}
\centering
  \includegraphics[width=1\linewidth]{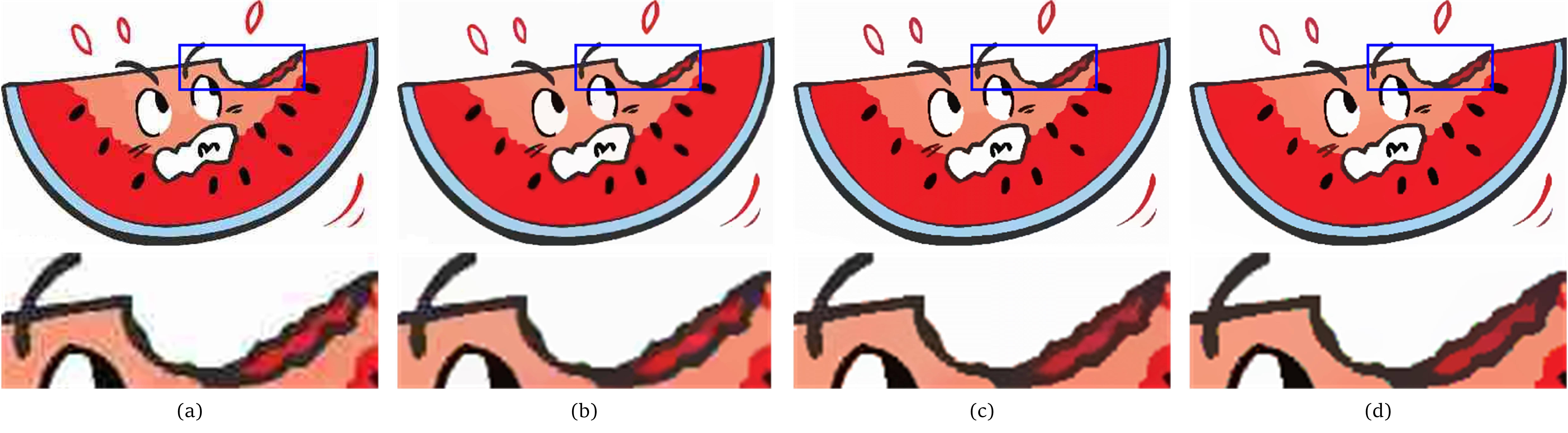}\\
  \caption{Clip-art JPEG compression artifacts removal. (a) Input compressed JPEG image. Result of (b) Xu et al.\cite{xu2011image}, (c) JWMF \cite{zhang2014100+} and (d) our method. Regions in blue boxes are highlighted.}\label{FigJPEGArtifactsRemove}
\end{figure*}
\begin{figure*}
\centering
  \includegraphics[width=1\linewidth]{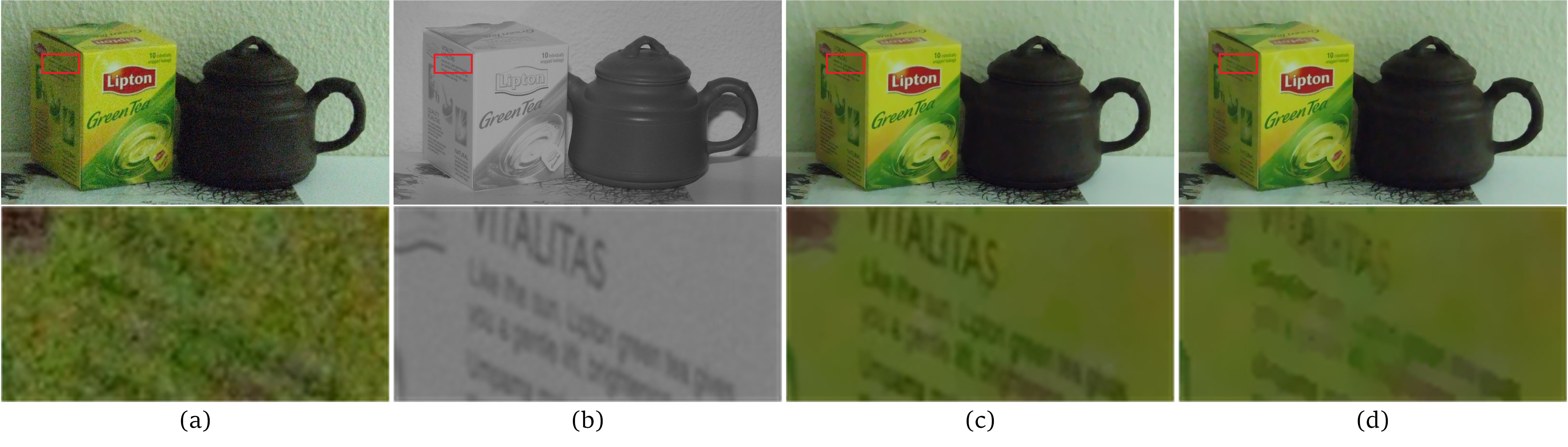}\\
  \caption{Our method cannot restore small details that exist in the guidance image but do not exist in the target image. Methods have strong texture copy ability can restore such details. (a) Input color image with noise. (b) Guidance NIR image. Result of (c) Shen et al. \cite{shen2015multispectral} and (d) our method. Regions in red boxes are highlighted.}\label{FigLimitation}
\end{figure*}

\subsection{Texture Smoothing}

For many cases, semantically meaningful structures are formed or appear over textured surfaces. Xu et al.\cite{xu2012structure} defined this kind of images as ``structure + texture`` images. Extracting these structures under the complication of texture patterns, which could be regular, near-regular, or irregular, is very challenging, but of great practical importance \cite{xu2012structure}. Inspired by the $L_2$ norm Total Variation (TV) \cite{aujol2006structure}, Xu et al. \cite{xu2012structure} proposed to perform texture smoothing with Relative Total Variation (RTV) which is a global method and can efficiently smooth textures while properly preserving structures. However, RTV can blur weak edges in some cases. We show an example in Fig.~\ref{FigTextureSmooth} (c). Recently, local methods based on bilateral filtering \cite{tomasi1998bilateral} were proposed to perform texture smoothing, for example, Rolling Guidance Filtering (RGF) \cite{zhang2014rolling}, bilateral texture filtering \cite{cho2014bilateral} etc. These methods usually cannot complete smooth regions of textures as global methods do. An example result of RGF is illustrated in Fig.~\ref{FigTextureSmooth}(b). As illustrated before in Fig.~\ref{FigNoiseSmoothingIllustration} and Fig.~\ref{FigGuideWeightAdvantage}, our method has strong noise/texture smoothing property while preserving sharp edges. We pre-smooth input images with a $3\times3$ median filter to smooth very sharp edges in textures. Our texture smoothing result is shown in Fig.~\ref{FigTextureSmooth}(d) where textures are well smoothed while edges are properly preserved.

\subsection{Clip-art Compression Artifacts Removal}

Clip-art/carton images are quite piece-wise smooth with sharp edges. When compressed in JPEG format, clear compression artifacts will occur around edges. General denoising approaches do not suit this application as the compression artifacts are strongly correlated with edges. To restore degraded images, Wang et al. \cite{wang2006deringing} proposed an image analogies approach that required prior knowledge and a training process. Recently, there are also general smoothing methods applied to perform clip-art compression artifacts removal such as the gradient $L_0$ norm optimization approach proposed by Xu et al. \cite{xu2011image} and the Joint Weighted Median Filtering (JWMF) proposed by Zhang et al. \cite{zhang2014100+}. The gradient $L_0$ norm optimization \cite{xu2011image} cannot completely remove strong compression artifacts as illustrated in Fig.~\ref{FigJPEGArtifactsRemove}(b). There also artifacts remaining in the result of JWMF in Fig.~\ref{FigJPEGArtifactsRemove}(c). Our result is illustrated in Fig.~\ref{FigJPEGArtifactsRemove}(d) with artifacts better removed while edges are properly preserved.

\section{Conclusion and Limitation}
\label{SecConclusion}

In this paper, we propose a general framework named Robust Guided Image Filtering (RGIF). RGIF can be applied to both dual-image tasks and single-image tasks. For dual-image tasks, it is robust against not only the heavy noise in the target image but also the structure inconsistency between the guidance image and the target image, which is implemented through the special structure of the data term and the smoothness term respectively. We believe the proposed work shows new insights in handling the challenging problems in guided image filtering. The first one is that we can simultaneously smooth the target image and perform guided image filtering instead of separating them as two steps. We have shown this can lead to gain better performance. The second one is that we can make use of the property of both the guidance image and the target image to handle the structure inconsistency other than using additional methods to assist the filtering process such as shadow detection \cite{ petschnigg2004digital, eisemann2004flash, shen2015multispectral}. We demonstrate that this can handle more situations other than the simple shadow problem. For single-image tasks, our method also shows promising performance in strong smoothing property while properly preserving edges. It is capable of handling several challenging applications in avoiding halos, gradient reversals and well preserving edges with noise/texture properly smoothed.

Texture copy should be avoided in most cases in guided image filtering. However, if the latent structure, i.e., the structure that is invisible, of the target image is consistent with that of the guidance image, texture copy can enhance the image quality instead of causing artifacts. We show such an example in Fig.~\ref{FigLimitation}(a) and (b) where characters that are seldom visible in the noisy color image appear in the flash NIR image. These characters are seldom restored in our result in Fig.~\ref{FigLimitation}(d) because our method suppresses texture copy. However, they are clearly restored in the result by Shen et al. \cite{shen2015multispectral} in Fig.~\ref{FigLimitation}(c). In this case, the ability of suppressing texture copy of our method should be regarded as its limitation.

\bibliographystyle{IEEEtran}

\bibliography{egbib}

\end{document}